\documentclass[conference]{IEEEtran}
\IEEEoverridecommandlockouts

\usepackage{cite}
\usepackage{amsmath,amssymb,amsfonts}
\usepackage{algorithmic}
\usepackage{graphicx}
\usepackage{textcomp}
\usepackage{xcolor}
\usepackage{CJKutf8}
\usepackage{amsmath}
\usepackage{booktabs}

\def\BibTeX{{\rm B\kern-.05em{\sc i\kern-.025em b}\kern-.08em
    T\kern-.1667em\lower.7ex\hbox{E}\kern-.125emX}}
\begin{document}

\title{Collaborative Attention and Consistent-Guided Fusion of MRI and PET for Alzheimer’s Disease Diagnosis\\
{\footnotesize \textsuperscript{}}
\thanks{}
}

\author{\IEEEauthorblockN{Delin Ma}
\IEEEauthorblockA{\textit{School of Software} \\
\textit{Yunnan University}\\
Kunming, China \\
Delinm176@gmail.com}
\and
\IEEEauthorblockN{Menghui Zhou}
\IEEEauthorblockA{\textit{Department of Computer Science} \\
\textit{The University of Sheffield}\\
Sheffield, UK \\
menghui.zhou@sheffield.ac.uk}
\and
\IEEEauthorblockN{Jun Qi}
\IEEEauthorblockA{\textit{Department of Computing} \\
\textit{Xian JiaoTong-Liverpool University}\\
Suzhou, China \\
Jun.Qi@xjtlu.edu.cn}
\and
\IEEEauthorblockN{Yun Yang}
\IEEEauthorblockA{\textit{School of Software} \\
\textit{Yunnan University}\\
Kunming, China \\
yangyun@ynu.edu.cn}
\and

\IEEEauthorblockN{Po Yang*}
\IEEEauthorblockA{\textit{Department of Computer Science} \\
\textit{The University of Sheffield}\\
Sheffield, UK \\
po.yang@sheffield.ac.uk}
}
\maketitle

\begin{abstract}
Alzheimer’s disease (AD) is the most prevalent form of dementia, and its early diagnosis is essential for slowing disease progression. Recent studies on multimodal neuroimaging fusion using MRI and PET have achieved promising results by integrating multi-scale complementary features. However, most existing approaches primarily emphasize cross-modal complementarity while overlooking the diagnostic importance of modality-specific features. In addition, the inherent distributional differences between modalities often lead to biased and noisy representations, degrading classification performance. To address these challenges, we propose a Collaborative Attention and Consistent-Guided Fusion framework for MRI and PET based AD diagnosis. The proposed model introduces a learnable parameter representation (LPR) block to compensate for missing modality information, followed by a shared encoder and modality-independent encoders to preserve both shared and specific representations. Furthermore, a consistency-guided mechanism is employed to explicitly align the latent distributions across modalities. Experimental results on the ADNI dataset demonstrate that our method achieves superior diagnostic performance compared with existing fusion strategies.
\end{abstract}


\section{Introduction}

Alzheimer’s disease (AD) is a progressive and irreversible neurodegenerative disorder and the leading cause of dementia, which has rapidly emerged as a major public health challenge and imposes a substantial societal burden. By 2050, the prevalence of dementia is projected to double in Europe and increase more than twofold worldwide, underscoring its profound clinical significance \cite{scheltens2021alzheimer}. Mild cognitive impairment (MCI), the prodromal stage of AD, is characterized by measurable declines in cognitive function without significant disruption of daily activities \cite{portet2006mild}. Although the precise etiology of AD remains unclear and no curative treatment currently exists once symptoms manifest, early diagnosis is crucial, as timely interventions can delay disease progression and improve patient outcomes \cite{scheltens2021alzheimer}. Neuroimaging techniques such as structural magnetic resonance imaging (sMRI) and positron emission tomography (PET) provide non-invasive approaches for detecting brain alterations associated with AD. Specifically, sMRI highlights structural abnormalities such as brain atrophy, whereas PET (e.g., FDG-PET) reflects functional impairments, including disrupted glucose metabolism \cite{damulina2020cross, nordberg2010use}. In recent years, machine learning and deep learning methods have been extensively applied to multimodal neuroimaging studies for AD diagnosis \cite{abdelaziz2022fusing, li20223,10495318,chang2024informative}. These studies consistently demonstrate that multimodal integration provides complementary perspectives, yielding more discriminative disease representations than single-modality approaches \cite{zhou2025enhanced, qiu20243d}. Nevertheless, due to the fundamentally different imaging mechanisms of sMRI and PET, their physical properties and data distributions exhibit substantial discrepancies. Direct fusion of these modalities often introduces noise and bias into the integrated features, which in turn degrades the performance of downstream classification tasks.

To address these challenges, recent studies have highlighted the importance of aligning multimodal neuroimaging data within a shared feature space. For instance, contrastive learning techniques have been employed to learn unified representations, thereby enhancing the integration of complementary information across modalities \cite{yu2025weighted, chen2022contrastive}. Alternatively, generative adversarial network (GAN)-based strategies have been proposed, where one modality is used to generate the other, and the synthesized images are subsequently utilized for AD detection \cite{chen2025joint}. While these approaches show promise in mitigating modality discrepancies, their effectiveness is often limited by the quality of contrastive learning sample pairs or the fidelity of the generated images, which can adversely affect downstream diagnostic performance. HybridCA-Net \cite{zhou2025enhanced} introduces a self-supervised consistency mechanism that employs the mean squared error (MSE) to quantify feature alignment. This mechanism is grounded in the theoretical assumption that effective multimodal fusion necessitates the establishment of a shared representation space. However, owing to the statistical nature of MSE, disparities in modality variance can lead the optimization process to be dominated by the high-variance modality. Furthermore, since inter-modal discrepancies are often nonlinear, relying solely on MSE may inadvertently distort the semantic integrity of individual modalities. To overcome these limitations, we employ a Feature-level Cross-Correlation strategy to more effectively capture the intrinsic correlations between modality-specific features.
In addition, effectively integrating modality-shared and modality-specific features remains a challenging task. Achieving comprehensive representations requires not only capturing complementary structural and functional information across modalities, but also preserving modality-specific discriminative patterns and modeling their nonlinear relationships. Simple strategies, such as feature concatenation or late decision-level fusion, often fail to capture these intricate cross-modal interactions, thereby limiting the full exploitation of multimodal complementarity.


To address this challenge, Wang et al. \cite{wang2024alzheimer} proposed a modality-specific and shared representation learning (MSSRL) framework, which integrates intra-modality specific information with inter-modality shared features. While effective in principle, this approach relies exclusively on convolutional neural networks for feature extraction, limiting its ability to capture comprehensive global representations and thereby constraining the richness of the learned multimodal features.



To address the aforementioned challenges, we propose a novel framework that leverages both MRI and PET modalities as input for AD diagnosis. Specifically, a feature extraction (FE) module is designed to learn both modality-specific and shared representations. By incorporating spatial, channel, and pixel attention mechanisms, the FE module effectively highlights salient regions, thereby enhancing the representational capacity of the model. Building on this, a cross-modal consistent feature enhancement (CCFE) module is introduced to capture shared features through parameter-sharing encoders and consistency constraints. Finally, the modality-specific and shared representations are integrated to provide a more comprehensive and discriminative perspective for disease prediction.

We conducted extensive experiments on the publicly available Alzheimer’s Disease Neuroimaging Initiative (ADNI) dataset \cite{Durazzo_Mattsson_Weiner_2014}, and the results demonstrate that our method outperforms existing approaches.


The main contributions of this work are as follows:
\begin{itemize}
\item To address the challenge of effectively integrating heterogeneous multimodal information in Alzheimer’s disease (AD) diagnosis, we develop a unified multimodal framework that combines modality-specific feature extraction (FE), cross-modal consistent feature enhancement (CCFE), and shared–specific feature fusion (SSFF). This architecture enables complementary exploitation of structural and functional imaging information, facilitating more comprehensive and discriminative representations for AD classification.

\item To overcome the limitations of conventional feature extractors in capturing subtle pathological patterns under low contrast or complex brain morphology, we design a triple-channel attention (TCA) module with residual connections. By jointly modeling spatial, channel, and pixel-level dependencies, this module enhances feature discriminability and improves sensitivity to fine-grained structural and metabolic variations.
\item We tackle the challenge of modality heterogeneity by introducing a cross-modal consistent feature enhancement (CCFE) mechanism with a shared-weight encoder and a learnable parameter representation (LPR) block, enabling adaptive information exchange and improving cross-modal feature consistency.

\end{itemize}

The remainder of this paper is organized as follows. Section II reviews the related work. Section III provides a detailed description of the proposed method. Section IV presents the experimental results and analysis. Finally, Section V concludes the paper.

\begin{figure*}[t]
  \centering
  \includegraphics[width=0.9\textwidth]{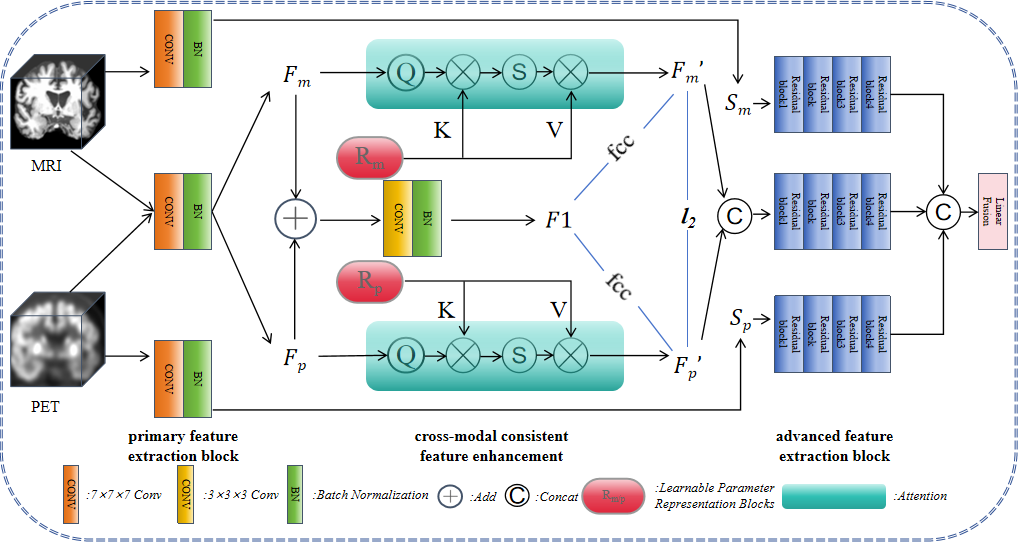}
  \caption{The proposed framework for Alzheimer’s disease diagnosis. The network first extracts modality-specific and shared representations through dedicated/shared feature extraction (FE) blocks. Subsequently, the Cross-modality Consistent Feature Enhancement (CCFE) module employs a learnable parameter representation (LPR) block to achieve MRI and PET information complementation at the feature level, while imposing a feature-level cross-correlation (FCC) consistency constraint to reduce modality discrepancies. Finally, the Shared–Specific Feature Fusion (SSFF) combines the enhanced shared features with the original modality-specific features and feeds them into a classifier for disease prediction.}
  \label{fig:frame}
\end{figure*}

\section{Related Work}
\begin{CJK}{UTF8}{gbsn}

In recent years, multimodal neuroimaging fusion for early AD detection has been extensively studied, and existing approaches can be broadly categorized into three levels: early fusion, feature-level fusion, and decision-level fusion.

Early fusion combines data from different modalities prior to processing, directly integrating pixel-level information from source images at the input stage. This type of fusion can enhance the visualization of complementary anatomical and functional details, but it requires highly accurate registration and is often sensitive to noise, which may introduce information bias and reduce model reliability. Several studies have adopted early fusion to improve diagnostic accuracy. For example, Z. Kong et al. combined MRI and PET images by summation and applied a single-channel network for subject classification, while El et al. \cite{el2020multimodal} proposed an integrated deep learning model based on stacked convolutional neural networks and bidirectional long short-term memory networks to extract statistical features from time series and improve performance.

Feature-level fusion extracts and combines features from different modalities, typically by generating feature vectors from each modality and then integrating them through concatenation, weighting, or cross-attention mechanisms. Although this strategy allows for richer representation, many existing methods are linear and thus struggle to capture the complex nonlinear interactions between modalities. Consequently, effectively leveraging the combined features has become a central research focus. For instance, Choulhury et al. \cite{choudhury2024coupled} encoded MRI and PET images into a shared latent space using autoencoders, and the latent feature vectors were then fused. Similarly, Zhang et al. \cite{zhang2021alzheimer} introduced a classification framework based on multimodal neuroimaging that incorporated embedded feature selection and fusion, adding a norm regularization term to integrate different kernel functions and exploit complementary information. Moreover, Abdelaziz et al. \cite{abdelaziz2025multi} employed a fusion mechanism with multi-head self-attention and cross-attention to capture both global relationships among features and the contribution of each modality to the other.

Decision-level fusion (also known as late fusion) combines the outputs of individual classifiers or algorithms trained on each modality \cite{chen2025joint}. Fusion strategies such as majority voting or weighted averaging are employed to aggregate the independent predictions. This approach does not rely on modality alignment and allows the use of modality-specific deep learning models for feature extraction. However, it fails to explicitly model cross-modal interactions and thus cannot fully exploit the potential of multimodal data.

Overall, most existing multimodal neuroimaging fusion methods focus on either modality-specific information or modality-shared information in isolation, neglecting their integration. Moreover, they often tolerate the inherent discrepancies introduced by multimodal heterogeneity. In this study, we introduce a learnable cross-modal consistent feature enhancement module to reduce modality discrepancies. By subsequently integrating modality-specific information, our model captures cross-modal interactions from a more comprehensive perspective, thereby improving prediction accuracy.
\end{CJK}
\section{Methodes}
\subsection{Overview}
\begin{CJK}{UTF8}{gbsn}
As illustrated in Figure \ref{fig:frame}, our proposed model consists of three main components: modality-specific feature extraction (FE), cross-modal consistent feature enhancement (CCFE), and shared–specific feature fusion (SSFF). The FE module extracts both modality-specific and shared multi-scale features from pre-registered image pairs. To enhance cross-modal feature representation, a shared-weight encoder is employed to simultaneously extract features from both modalities. Within the CCFE module, features are enhanced by incorporating information from the other modality, which effectively reduces discrepancies between MRI and PET features while promoting the establishment of cross-modal feature correspondences. Finally, the SSFF module integrates modality-specific and shared features, compensating for the modality-specific information that may be overlooked during shared feature learning. Further details of each component are provided in the following sections.
\end{CJK}

\subsection{Feature Extraction}
Leveraging the strong feature extraction capability of 3D ResNet, our FE module retains the residual connections while incorporating spatial, channel, and pixel attention to construct a feature encoder that effectively exploits the synergistic spatial–channel–pixel attention mechanism. This design aims to enhance the perception of challenging features in low-contrast or morphologically variable samples. By integrating attention mechanisms at multiple scales, the model can better focus on salient regions. Additionally, to strengthen the modeling of spatial relationships, deep convolutional blocks are employed within the spatial attention component.

Specifically, the FE module consists of three feature extraction components: a primary feature extraction block (PFE), an advanced feature extraction block (AFE), and a triple collaborative attention (TCA) module. The PFE extracts initial features using a 3D ResNet block with reduced channel dimensions and comprises a 7×7×7 convolutional layer with stride 2, an InstanceNorm layer, a ReLU activation, and a max-pooling layer. The AFE consists of four 3D ResNet basic blocks with the channel number doubling at each layer, ranging from 16 input channels to 128 output channels.

The TCA module integrates Channel Attention (CA), Spatial Attention (SA), and Pixel Attention (PA) to guide the network’s focus toward critical regions. In CA, global average pooling and max pooling compress the input features spatially to capture global context, followed by two 1×1 convolution layers with ReLU activation to generate a channel attention map, which is normalized via a Sigmoid function to determine the importance of each channel. The final channel attention map is obtained by weighting and combining the average- and max-pooled results. SA employs a series of depthwise separable convolutions with kernels of sizes 5×5, 1×7, 7×1, 1×11, 11×1, 1×21, and 21×1 to capture multi-scale spatial information. The resulting spatial feature maps are summed and fused via a 1×1 convolution to produce the final spatial attention map. In PA, during forward propagation, two convolution layers are applied to generate a pixel attention map, which is normalized with a Sigmoid function. The output is then obtained by multiplying the original input with the pixel attention map, emphasizing important pixel-level regions.

For the purpose of subsequent feature integration, two independent feature extractors, $FE_{m}$ and $FE_{p}$, are employed to extract features from MRI and PET images, respectively, along with a classifier for separate training. Additionally, two shared-weight feature extractors, $PFE_{sh}$ and $AFE_{sh}$, are used to extract features from both modalities simultaneously.

\begin{equation}
SF_{mri/pet}=FE_{m/p}(M/P)\label{eq}
\end{equation}

\begin{equation}
F_{m/p}=FE_{sh}(M/P)\label{eq}
\end{equation}

\subsection{Cross-modal Consistent Feature Enhancement}

Inspired by \cite{li2025mulfs}, the CCFE module consists of a feature fusion (FF) component and two learnable parameter representation blocks,$LPR_m$and$LPR_p$.The FF is designed to efficiently perform low-cost fusion of feature maps from different modalities. As illustrated in Figure \ref{fig:frame}, the initial alignment and integration of cross-modal information is achieved via element-wise addition:
\begin{equation}
F=F_{m}+F_{p}\label{eq}
\end{equation}
Here,$F_{m}$ and $F_{p}$represent the features extracted from MRI and PET images by the PFE module. These features are subsequently processed through convolution, normalization, and activation operations to further capture contextual information and adjust the feature distributions:

\begin{equation}
F_1=\sigma(BN(Conv(F)))\label{eq}
\end{equation}
Since $F_1$ contains both modality-specific and modality-shared information, we use the fused feature $F_1$ to supervise the single-modality feature extraction process. Specifically, the LPR blocks are employed to enhance cross-modal consistent feature and compensate for missing modality-specific information. The goal of LPR is to preserve the unique characteristics of either the MRI or PET modality as much as possible, while ensuring that features extracted from one modality also incorporate information from the other. This facilitates the extraction of modality-consistent features from single-modality images.

Let $R_{m}$ and $R_{p}$ denote the learnable parameter representations.$W^Q_{m/p}$,$W^K_{p/m}$ and $W^V_{p/m}$ represent linear mappings applied to $F_{m/p}$ and $R_{m/p}$,Specifically,we have $Q_{m/p}=W_{m/p}^QF_{m/p}$,$K_{p/m}=W^k_{p/m}R_{p/m}$,$V_{p/m}=W_{p/m}^VR_{p/m}$,With the aid of $R_{m/p}$,the enhanced single-modality MRI/PET features can be expressed as:
\begin{equation}
F_{m/p}^{\prime}=softmax(\tfrac{Q_{m/p}K_{p/m}^{T}}{\sqrt{d}})V_{p/m}\label{eq}
\end{equation}
The resulting features,$F_{m}^{\prime}$ and $F_{p}^{\prime}$,supplement the missing information from the other modality. This step reduces the discrepancies between the two modality-specific features, thereby increasing the information content of $F_{m/p}^{\prime}$.To ensure consistency between the enhanced features, we employ Feature-level Cross-Correlation (FCC)\cite{sun2021loftr} to measure the correlation between them
\begin{equation}
\begin{aligned}
&FCC(X,Y)=\\
&\frac{ \underset{i,j,k}{\sum} \left( X \left( i,j,k \right)-X \left( k \right) \right) \left( Y \left( i,j,k \right)-Y \left( k \right) \right)}{ \sqrt{ \underset{i,j,k}{\sum}  \left( X \left( i,j,k \right)-X \left( k \right) \right)^{2}} \sqrt{  \underset{i,j,k}{\sum}  \left( Y \left( i,j,k \right)-Y \left( k \right) \right)^{2}}}
\end{aligned}
\end{equation}
$R_{m}$ and $R_{p}$ can be learned using the following loss function:
\begin{equation}
l_{\text{consi}} = -FCC(R_{\text{m}}, F_1) - FCC(R_{\text{p}}, F_1)
\end{equation}
To ensure consistency and balance between $R_{m}$ and $R_{p}$ as well as between $R_{m/p}$ and $F1$,we introduce a mean squared error (MSE) constraint:
\begin{equation}
l_{\mathsf{mse}} = \frac{||fm^{ \prime}-fp^{\prime}||^2_F}{HWD}
\end{equation}
The loss $L_{mse}$ encourages the projections of different modalities to remain numerically close, preventing excessively large feature discrepancies caused by modality heterogeneity.

\begin{table*}

\caption{Performance Comparison of Different Methods}
\label{tab:db}
\centering
\begin{tabular}{ccccccccc}
\hline
Task & Method &Year&Modality& ACC (\%) & AUC (\%) & PRE (\%)& SPE (\%)& SEN (\%) \\
\hline
& MSSRL       &2024&MRI+PET  & 93.79   & 97.54    & 93.98   &  94.62  & 92.86       \\
&3DResNet-18  &2018&MRI+PET  &88.02$\pm1.95$   & 95.30$\pm1.04$&86.11$\pm5.46$& 90.84$\pm3.12$        &90.00$\pm4.62$ \\
& Zhao et al.   &2022&MRI+PET  &90.11$\pm$2.61   &95.79$\pm2.16$    & 85.73$\pm7.18$   & 91.30$\pm4.32$   &87.44$\pm7.99$\\

ADvCN&MMSDL        &2025&MRI+PET  &87.68$\pm1.28$  & 96.68$\pm0.84$    & 88.43$\pm1.06$   & 84.73$\pm6.61$   &87.84$\pm1.24$        \\
&MDL-Net &2024&GM+WM+PET&88.81$\pm1.21$& 96.88$\pm1.69$&81.19$\pm4.07$ &88.28$\pm3.28$   &89.79$\pm3.15$  \\
&HybridCA-Net&2025&MRI+PET& 91.25$\pm2.81$  &96.30$\pm1.62$   & 93.89$\pm6.50$   & 95.36$\pm7.71$  &83.36$\pm7.81$        \\
&ours       &-&MRI+PET&\textbf{94.40$\pm$2.28}&\textbf{ 97.16$\pm$1.96}&\textbf{96.56$\pm$3.23}&\textbf{97.37$\pm$3.46}& 88.54$\pm$8.31 \\
\hline
& MSSRL       &2024&MRI+PET  & 73.55   & 76.34    & 73.55  &  70.97  & 74.86       \\
&3DResNet-18  &2018&MRI+PET  & 65.14$\pm4.23$& 77.10$\pm2.19$& 84.05$\pm6.70$&72.06$\pm22.94$         &60.87$\pm15.32$              \\
& Zhao et al.  &2022&MRI+PET  &67.84$\pm4.83$ & 72.64$\pm4.44$  & 74.29$\pm7.22$   & 56.81$\pm19.96$  & 74.65$\pm11.73$\\

CNvMCI &MDL-Net  &2024&GM+WM+PET&76.11$\pm3.26$ &83.76$\pm8.86$ &72.41$\pm5.98$    &82.84$\pm8.87$  & 65.53$\pm15.02$      \\
&HybridCA-Net      &2024&MRI+PET&70.74$\pm2.55$ &75.87$\pm6.30$ &72.17$\pm6.51$    &62.43$\pm14.45$   & 77.37$\pm10.14$     \\
&ours       &-&MRI+PET&73.07$\pm$4.39&77.37$\pm$4.21&64.25$\pm$10.60&73.92$\pm$6.77& 72.35$\pm$6.76 \\
\hline

&3DResNet-18  &2018&MRI+PET  &71.31$\pm2.83$ & 77.55$\pm4.17$& 83.15$\pm5.27$& 60.50$\pm24.35$  &73.88$\pm11.32$   \\
& Zhao et al.   &2022&MRI+PET  & 74.58$\pm4.16$   & 80.72$\pm3.52$   & 82.55$\pm6.73$    & 59.60$\pm15.23$   & 81.56$\pm11.17$\\
ADvMCI&MMSDL        &2025&MRI+PET  & 77.34$\pm2.05$   & 84.06$\pm0.47$    & 74.64$\pm1.96$  & 66.55$\pm11.58$   & 74.42$\pm3.49$  \\
&MDL-Net      &2024&GM+WM+PET&75.66$\pm3.06$ &82.22$\pm0.50$   & 82.28$\pm1.86$  &79.13$\pm5.44$  &73.00$\pm$9.15    \\
&HybridCA-Net&2025&MRI+PET&77.95$\pm3.56$ &81.65$\pm4.48$&84.53$\pm6.24$  &63.54$\pm13.35$   & 84.29$\pm6.93$    \\
&ours      &-&MRI+PET&\textbf{80.07$\pm$5.64} &82.54$\pm$7.55 &83.33$\pm$5.82 &\textbf{80.50$\pm$7.99} &79.80$\pm$11.4  \\
\hline
\end{tabular}

\end{table*}

\subsection{Specific versus Shared Fusion }
\begin{CJK}{UTF8}{gbsn}
To comprehensively integrate modality-specific and modality-shared features, we leverage MRI and PET feature extractors $FE_{m}$ and $FE_{p}$, independently pre-trained on single-modality classification tasks, to generate modality-specific features, $S_m$ and $S_p$ for MRI and PET images, respectively. This step ensures that each modality retains its discriminative information. Simultaneously, $(F_{m}^{\prime}, F_{p}^{\prime})$ are fed into the shared feature extraction network $AFE_{sh}$ to obtain the cross-modal shared feature $F_{sh}$. The shared features capture common information between MRI and PET modalities. Subsequently, the shared features $F_{sh}$ are concatenated with the modality-specific features $S_m$ and $S_p$ to form a joint feature vector $F_{joint} = [F_{sh}, S F_m, S F_p]$. Finally, $F_{joint}$ is passed through a linear classification layer to produce the predicted label $y_{pred}$. The classification loss is computed using the cross-entropy loss:
\begin{equation}
l_c=-\sum_{i=1}^{N}y_{i} \log \left( y_{pred,i} \right)
\end{equation}
The overall loss of the proposed method is defined as follows:
\begin{equation}
l_{total}=l_{consi}+\lambda   l_{mse}+l_c
\end{equation}
where $\lambda$ denotes the trade-off parameter, empirically set to 0.5 in this study.
\end{CJK}

\begin{table*}
\centering
\caption{Ablation Results in Three Classification Tasks}
\begin{tabular}{cccccc}

\hline
Task & Method & ACC (\%) & AUC (\%) & PRE (\%)& SEN (\%) \\
\hline
&ours-(CCFE+SSFF)&90.96&95.71&91.12&91.17\\
ADvCN&ours-(TCA+ssff)&89.26&94.59&90.43&89.73\\
&ours-SSFF&92.66&97.23&94.93&89.28\\
\hline

&ours-(CCFE+SSFF)&70.29&70.49&66.74&60.94\\
CNvMCI&ours-(TCA+ssff)&69.92&71.48&65.70&63.04\\
&ours-SSFF&70.65&75.12&67.20&67.29\\
\hline

&ours-(CCFE+SSFF)&77.15&80.16&73.55&72.06\\
ADvMCI&ours-(TCA+ssff)&73.78&74.90&73.58&60.90\\
&ours-SSFF&76.11&81.95&78.44&83.33\\

\hline
\label{tab:ablation}
\end{tabular}
\end{table*}
\section{Experiments And Results}
\begin{table}
\centering

\caption{Simplified Demographic Information from Two Datasets}
\begin{tabular}{ccccc}
\hline
Dataset&Cat.&M/F&Age&Edu\\
\hline
&CN&55/38&76$\pm$5&16$\pm$3\\
ADNI-1&AD&48/36&76$\pm$7&14$\pm$3\\
&MCI&121/62&75$\pm$7&16$\pm$3\\
\hline
&CN&114/131&73$\pm$6&17$\pm$3\\
ADNI-2&AD&78/58&74$\pm$8&16$\pm$3\\
&MCI&182/146&72$\pm$7&16$\pm$3\\
\hline

\label{tab:data}
\end{tabular}
\end{table}
\subsection{Materials and Preprocessing}
The dataset used in this study was selected from ADNI-1 and ADNI-2, including all subjects who had both MRI and PET scans available (see Table \ref{tab:data}). Subjects were categorized into three groups: AD, CN, and MCI. Data preprocessing was performed using FreeSurfer \cite{fischl2012freesurfer} and SPM12, which included skull stripping, normalization, registration, and segmentation of gray matter and white matter. Each MRI image was affine-transformed to the standard MNI152 template using SPM, and the corresponding PET image was adjusted using the same parameters to ensure anatomical alignment between MRI and PET scans. After preprocessing, during model training, all images were resized to 128×128×128 with voxel dimensions of 1.422, 1.704, 1.422 mm, respectively, to fit the network architecture and reduce the number of model parameters.

\subsection{Experimental Settings and Evaluation}

To compare the proposed method with other multimodal neuroimaging-based AD classification approaches, all experiments were conducted on the same dataset with consistent network hyperparameters. For all methods, the learning rate was set to 0.0001, the total number of iterations was 40, and the classification loss function was cross-entropy, optimized using the Adam optimizer. Models were trained on an NVIDIA GeForce RTX 4080 SUPER GPU with 16 GB of memory.

Considering the characteristics of AD at different stages, we conducted three binary classification tasks: Alzheimer's disease (AD) vs. cognitively normal (CN), mild cognitive impairment (MCI) vs. CN, and AD vs. MCI. In all tasks, the batch size was set to 4. After preprocessing the raw data, ten-fold cross-validation was applied for model training and testing. Models were evaluated using multiple classification metrics, including accuracy (ACC), specificity (SPE), sensitivity (SEN), precision (PRE), and area under the curve (AUC). Additionally, after each training step, the corresponding test results were recorded. Once the model stabilized, the final evaluation metrics were obtained by averaging all classification results across folds. The results are reported as the mean and standard deviation over the ten folds, providing a robust assessment of the effectiveness of the proposed method.

\subsection{Comparison with State-of-the-art Methods}

To demonstrate the superiority of the proposed method, we evaluate it on the ADNI1 and ADNI2 datasets and compare it with other multimodal neuroimaging-based AD detection approaches under identical experimental conditions. These methods include:
\begin{itemize}
    \item 3DResNet-18 \cite{he2016deep}: A classical deep learning classification network based on residual connections that effectively preserves the original image features. Since our proposed method is built upon its 3D version, it is included as a baseline for comparison.
    \item MSSRL \cite{wang2024alzheimer}: A multimodal specific–shared representation learning framework designed to assist Alzheimer’s disease diagnosis, which aims to integrate modality-specific information with cross-modal shared information.
    \item Zhao et al. \cite{kong2022multi}:Zhao et al. employed a novel fusion strategy and incorporated a sparse autoencoder into the network to enhance the feature representation capability.
    \item MMSDL \cite{abdelaziz2025multi}: A multi-scale, multimodal learning approach that leverages both local and global features to reduce noise and variability across scales and modalities, thereby improving predictive accuracy.
    \item MDL-Net \cite{qiu20243d}: Utilizes multifusion joint learning rather than a single fusion strategy, and incorporates a disease-induced perception learning module to enhance model interpretability.
    \item HybridCA-Net \cite{zhou2025enhanced}: HybridCA-Net extracts spatiotemporal dynamics from fMRI and fine-grained anatomical features from sMRI by deploying a spatiotemporal graph convolutional network and a 3D ResNet in parallel. A self-supervised consistency loss is introduced to explicitly align the distributional discrepancies between the two modalities in a shared latent space. In this study, PET is used as a substitute for fMRI.
\end{itemize}

It is worth noting that, except for MDL-Net, all other methods use MRI and PET images as inputs. MDL-Net, however, utilizes gray matter and white matter images derived from MRI segmentation, along with the corresponding PET images, as input. Due to differences in dataset samples and preprocessing procedures, direct comparison with existing studies is not feasible; nevertheless, we have endeavored to replicate their methodologies as closely as possible.

The experimental results are presented in Table \ref{tab:db}. Our proposed method achieves the highest accuracy in both the AD vs. CN and AD vs. MCI tasks, with values of 94.40\%$\pm$2.28\% and 80.07\%$\pm$5.64\%,respectively. 

\subsection{Loss Values Curves of Different Tasks}
\begin{figure}
    \centering
    \includegraphics[width=\linewidth]{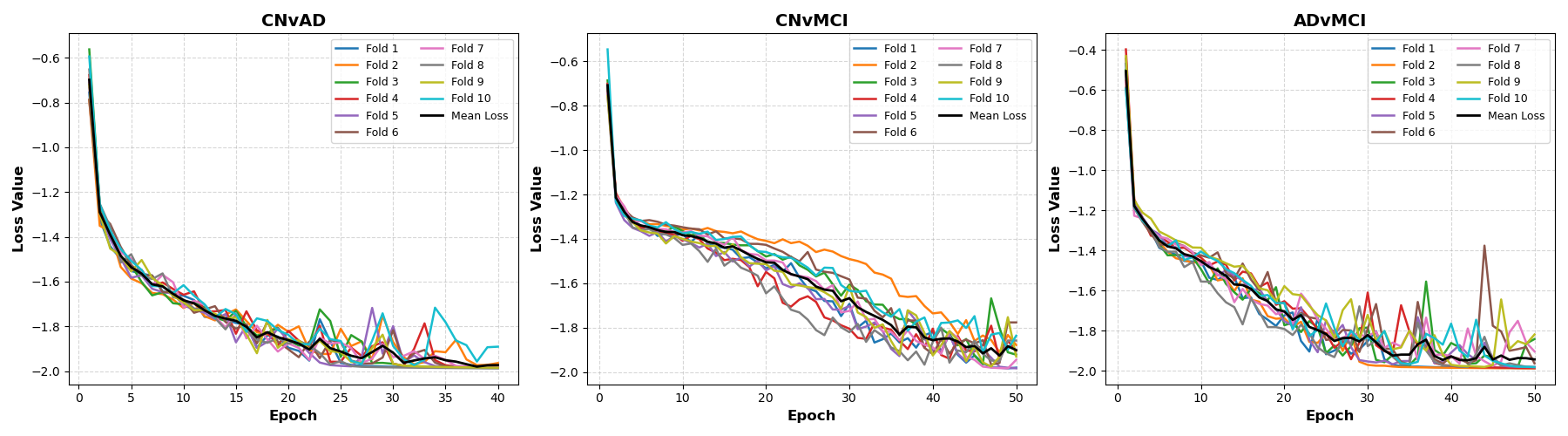}
    \caption{The loss value trends in the three classification tasks—(a) CN vs. AD, (b) CN vs. MCI, and (c) AD vs. MCI—illustrate the model’s training dynamics over the epochs. For each task, the loss curves of the 10 cross-validation folds show a consistent decline, indicating effective convergence. The mean loss curve, highlighted in black, represents the overall training trend, while the individual colored lines reflect fold-specific variations, demonstrating the model’s robustness and stability across different data splits.}
    \label{fig:placeholder}
\end{figure}
As shown in Figure \ref{fig:placeholder}, in the third fold of each task, convergence requires more iterations, which may be due to the presence of hard-to-distinguish samples; however, the overall trend of the loss remains consistent. Notably, in the latter two tasks, the rate of loss reduction is significantly lower than in the first task, particularly in the CN vs. MCI task. This reflects the substantial overlap in data features between MCI and CN, whereas the neurodegenerative changes in AD patients are more pronounced. Consequently, the differences between AD and CN or AD and MCI are more distinct, enabling the model to capture these differences more rapidly and achieve faster convergence.
\subsection{Ablation Study}
To validate the effectiveness of each component in the proposed model, we conducted ablation experiments on the same dataset. The results are presented in Table \ref{tab:ablation}. When individual modules were used alone or removed, the model’s performance decreased to varying degrees. Specifically, using only the TCA module, the accuracy for the AD vs. CN classification task reached 90.96\%, likely due to its strong capability to capture morphological changes in the brain. In contrast, using only the CCFE module led to a minor improvement in performance, as enhancing consistent feature between modalities via the shared encoder alone was insufficient to substantially increase the amount of captured information. Notably, when both modules were employed simultaneously, the model’s performance approached the optimal level, demonstrating the complementary benefits of TCA and CCFE in the proposed architecture.

\section{Conclusion}
In this study, we propose a multimodal neuroimaging fusion framework for AD diagnosis, which integrates sMRI and PET images to achieve accurate classification. To enhance the model’s ability to capture challenging features, we first design a Triple-Collaborative Attention (TCA) feature extractor that jointly models spatial, channel, and pixel-wise dependencies. Next, a Cross-modal consistent feature Enhancement (CCFE) module is introduced, employing learnable parameter representations to transfer complementary information across modalities and using a consistency loss to further reduce cross-modal discrepancies. Finally, an enhanced Shared–Specific Fusion (SSFF) strategy integrates the improved shared features with reliable modality-specific cues, ensuring that the classifier can fully leverage both intra- and cross-modal discriminative information.

We conducted experiments on 1,086 subjects from ADNI-1 and ADNI-2. The results demonstrate that the proposed method achieves the highest accuracies of 94.40\% for CN vs. AD and 80.07\% for AD vs. MCI. Comparisons with state-of-the-art methods indicate that our approach outperforms existing best-performing models.

\section*{Acknowledgments}
This work was supported by the National Natural Science Foundation of China (62301452)
\bibliographystyle{IEEEtran}

\end{document}